\documentclass[
]{ceurart}

\usepackage{listings}
\usepackage{amsmath}
\usepackage{algorithm}
\usepackage{algpseudocode}
\begin{document}

\copyrightyear{2024}
\copyrightclause{Copyright for this paper by its authors.
  Use permitted under Creative Commons License Attribution 4.0
  International (CC BY 4.0).}

\conference{Workshop on Advanced AI Methods and Interfaces for Human-Centered Assistive and Rehabilitation Robotics (a Fit4MedRob event) - AIxIA 2024, November 25--28, 2024, Bolzano, Italy}

\title{Time is on my sight: 
scene graph filtering for dynamic environment perception in an LLM-driven robot}
\tnotemark[1]


\author[1, 3]{Simone Colombani}[
email=simone.colombani@studenti.unimi.it,
]

\author[2]{Luca Brini}[
email=l.brini@campus.unimib.it, 
]

\author[2]{Dimitri Ognibene}[
email=dimitri.ognibene@unimib.it,
]

\author[1]{Giuseppe Boccignone}[
email=giuseppe.boccignone@unimi.it,
]

\address[1]{University of Milan, Italy}
\address[2]{University of Milano-Bicocca, Milan, Italy}
\address[3]{Oversonic Robotics, Carate Brianza, Italy}

\begin{abstract} 
Robots are increasingly being used in dynamic environments like workplaces, hospitals, and homes. As a result, interactions with robots must be simple and intuitive, with robots' perception adapting efficiently to human-induced changes. 
\\
This paper presents a robot control architecture that addresses key challenges in human-robot interaction, with a particular focus on the dynamic creation and continuous update of the robot’s state representation. The architecture uses Large Language Models to integrate diverse information sources, including natural language commands, robotic skills representation, real-time  dynamic semantic mapping of the perceived scene. This enables flexible and adaptive robotic behavior in complex, dynamic environments.
\\
Traditional robotic systems often rely on static, pre-programmed instructions and settings, limiting their adaptability to dynamic environments and real-time collaboration. In contrast, this architecture uses LLMs to interpret complex, high-level instructions and generate actionable plans that enhance human-robot collaboration.
\\
At its core, the system’s Perception Module generates and continuously updates a semantic scene graph using RGB-D sensor data, providing a detailed and structured representation of the environment. A particle filter is employed to ensure accurate object localization in dynamic, real-world settings.
\\
The Planner Module leverages this up-to-date semantic map to break down high-level tasks into sub-tasks and link them to robotic skills such as navigation, object manipulation (e.g., PICK and PLACE), and movement (e.g., GOTO). \\
By combining real-time perception, state tracking, and LLM-driven communication and task planning, the architecture enhances adaptability, task efficiency, and human-robot collaboration in dynamic environments.
\end{abstract}



\maketitle
\section{Introduction}
Immediacy is crucial in assistive robotics  \cite{di2023personalized,lucignano2013dialogue, ognibene2019}. In a typical human-robot interaction scenario, users may provide commands in natural language, such as “\emph{Pick the blue bottle on the table and bring it to me}”.
To such aim, the use of Large Language Models (LLM) allows robots to interpret natural language requests and ``translate'' instructions into plans to achieve specific goals; yet, these models need to know the environment in which they operate so to generate accurate plans \cite{galindo2008robot}. 
The need for translation arises from the complexity of human language and the variability in instructions. Users may express commands differently or exploit ambiguous terms that the robot must comprehend. To address these challenges, robotic architectures must integrate natural language processing with environmental understanding.
\\
The chief concern of the work  is to exploit scene graphs as  semantic maps  providing a structured representation of spatial and semantic information of robot's environment. This enables LLMs to generate plans based on this information.
Indeed, via scene graphs robots can map the relationships between objects, their properties, and their spatial arrangements. 
\\
Here we address such limitations  by representing the environment as a  graph endowed with updatable semantics that language models can interpret. More precisely, the dynamics of the update is achieved via  particle filtering  to enhance the reliability and precision of real-time semantic mapping. The  model adopted (PSGTR) is lightweight and can be easily utilized, making it suitable for live applications and accessible even on less powerful hardware.
Using RoBee, the cognitive humanoid robot developed by Oversonic Robotics, the system dynamically updates the environment graph and replans in case of failure, overcoming challenges in long-term task planning.

\section{Related works}
A scene graph captures detailed scene semantics by explicitly modeling objects, their attributes, and the relationships between paired objects (e.g., “blue bottle on the table”) \cite{zhu2022scene}. 3D scene graphs \cite{armeni20193d} extend this concept to three-dimensional spaces, representing environments like houses or offices, where each piece of furniture, room, and object is a node. The edges between these nodes describe their relationships, such as a vase \textit{on} a table or a chair \textit{in front of} a sofa. 
\\
Recent works, such as 
\cite{gu2024conceptgraphs} and \cite{chang2023context} have  proposed to generate 3D scene graphs from RGB-D images, combining geometric and semantic information to create detailed environmental representations. Scene graphs have been widely used in computer vision and robotics to improve scene understanding, object detection, and task planning. For example, SayPlan \cite{rana2023sayplan} integrates 3D scene graphs and LLMs for task navigation and planning, performing semantic searches on the scene and instructions to create accurate plans, further refined through scenario simulations. DELTA \cite{liu2024delta} utilizes 3D scene graphs to generate PDDL files, employing multiple phases to prune irrelevant nodes and decompose long-term goals into manageable sub-goals, enhancing computational efficiency for execution with classical planners. SayNav \cite{rajvanshi2024saynav} constructs scene graphs incrementally for navigation in new environments, allowing the robot to generate dynamic and appropriate navigation plans in unexplored spaces by passing the scene graph to a LLM, thus facilitating effective movement and execution of user requests.
\\
In a crude summary, the main limitations of the above mentioned approaches to build environment representations lie in their reliance on computationally heavy vision-language models (VLMs) and computer vision models. Such models are not designed for precision and often demand significant resources, while  lacking the ability to be updated in real time, and thus limiting their practical application.

\section{Architecture}
Our system is based on two components:
\begin{itemize}
    \item \textbf{Perception Module}: it is responsible for sensing and interpreting the environment and building a semantic map in the form of a directed graph that integrates both geometric and semantic information. Its architecture is explained in detail below.
    
    \item \textbf{Planner Module}: it takes the information provided by the Perception Module to formulate plans and actions that allow the robot to perform specific tasks.
It is composed by the following:
\begin{itemize}
    \item Task Planner: Translates user requests, expressed in natural language, into high-level skills.
\item Skill Planner: Translates high-level skills into specific, low-level executable actions.
\item Executor: Executes the low-level actions generated by the Skill Planner.
\item Controller: Monitors the execution of actions and manages any errors or unexpected events
during the process.
\item Explainer: Interprets the reasons of execution failures by analyzing data received from the
Controller and provides suggestions to the Task Planner on how to adjust the plan.
\end{itemize}
\end{itemize}

These components interact to allow the robot to understand its environment and act accordingly to satisfy user requests. In what follows we specifically address the Perception Module while  details on the planner will be provided in a separate article.

\emph{Robot Hardware.}
The system was implemented using RoBee, the cognitive humanoid robot developed by Oversonic Robotics. RoBee, shown in Figure \ref{fig:semantic_map_example_and_robee}, stands 160 cm tall and weighs 60 kg. It features 32 degrees of freedom, and is equipped with cameras, microphones, and force sensors.

\subsection{Perception module}
The Perception Module is the component responsible for building a representation of the environment, which the robot can use for task planning.
The representation takes the form of a semantic map, a graph that integrates both geometric and semantic information about the environment. 
To generate the semantic map, the perception module uses data from various sensors. It requires RGB-D frames obtained from the camera  which are then processed using a scene graph generation model, such as PSGTR \cite{yang2022panoptic} to extract objects masks, label and relationships. Also it uses data on the camera position relative to the geometric map to determine the location of the objects identified by the model.
More formally, a Semantic Map is represented as a directed graph \( G_m = (V_m, E_m) \) where:
\begin{itemize}
    \item A node $v \in V_m$ can be one of the following types:
    \begin{itemize}
        \item Room node: Defines the different semantic areas of the environment, such as ``kitchen,'' ``living room,'' or ``bedroom.'' Each room node contains information about its geometric boundaries and the object nodes it contains;
        \item Object node: Represents physical objects in the environment, such as ``table,'' ``chair,'' or ``bottle.'' Each object node contains information about its 3D position, semantic category, dimensions, and other relevant properties:
    \end{itemize}
    \item An edge $e \in E_m$ can represent:
    \begin{itemize}
        \item The relationship between two objects;
        \item The connection between two rooms;
        \item The belonging of an object to one and only one room.
    \end{itemize}
\end{itemize}
The presence of room nodes is important because it facilitates the categorization of objects based on their respective rooms, which helps distinguish between objects with the same name and enhances the natural language description of the task, while room nodes enable the application of graph search algorithms for planning paths to objects. Room nodes are created based on the geometric map, while object nodes are generated following the steps explained below.

As to edges, more specifically:
\begin{itemize}
    \item \textbf{Edges between rooms} directly connect two rooms and facilitate navigation between them.
    \item \textbf{Edges between objects} represent the relationships between objects and are directed, the direction capturing the influence of one object on another; the label associated with each edge is derived from the inferences made by the PSGTR model.
\end{itemize}

\begin{figure*}
    \centering
    \begin{minipage}{0.7\textwidth}
        \includegraphics[width=\textwidth]{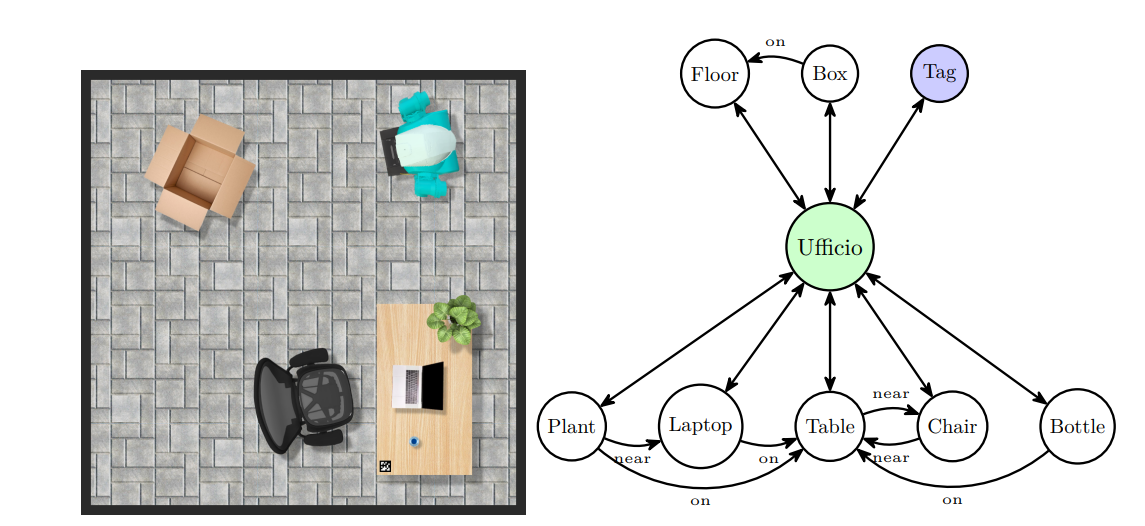}
    \end{minipage}\hfill
    \begin{minipage}{0.225\textwidth}
        \centering
    \includegraphics[width=\textwidth]{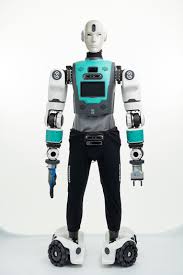}
    \end{minipage}
    \label{fig:semantic_map_example_and_robee}
    \caption{The figure on the left showcases an example of a semantic map in an office environment, while the image on the right shows RoBee, the humanoid robot developed by Oversonic Robotics.}
\end{figure*}

Figure \ref{fig:semantic_map_example_and_robee} shows an example of a semantic map of an office, built with the room node 'Office' (italian, 'Ufficio') and the object nodes connected to each other by relationships and linked to the room node.

\paragraph{Generating and updating the semantic map}
The scene graph generation process is based on the PSGTR model, a single-stage model built on the Transformer architecture \cite{10.5555/3295222.3295349}. This model generates a graph representation of a scene given its panoptic segmentation. PSGTR does not achieve the highest quality in panoptic segmentation compared to better models, but it provides reasonable inference times for real-time applications, taking about 400 ms to process a 480p image on a machine with access to an NVIDIA T4 GPU.
\\
The Perception Module uses the result of PSTGR  and builds the semantic map following the steps below:
\begin{enumerate}
    \item \textbf{Reading RGB-D frames}: 
    The video frames from the robot's cameras are sent to the model to be analyzed and used to generate the scene graph.    
    \item \textbf{Reading robot poses}: 
    To generate the scene and semantic map, it is necessary to know the robot's position relative to the geometric map, the camera's position relative to the map, and the camera's mounting position on the robot.    
    \item \textbf{Inference}: 
    Each received frame is processed by the model. Results are information about detected objects, such as labels and masks, and the relationships between them, such as relationship labels and associated probabilities.    
    \item \textbf{Graph construction}: 
    This step involves extracting data from the object returned by the model and computing values dependent on the robot system, such as the position of objects. At a finer level it consists of three sub-steps:
    \begin{enumerate}
        \item \textbf{Node construction}: 
        Classes and masks of detected objects are extracted. Next, the 3D position of each object is computed, starting in the pixel coordinate system, then transforming to the camera system, and finally to the robot's map coordinate system. Nodes for the semantic scene and the semantic map are instantiated using the appropriate 3D coordinates. 
        A distance-based filter is applied to prune objects that are too far from the robot to avoid issues with object detection and tracking.
        
        \item \textbf{Edge construction}: 
        Data about relationships between objects are extracted. For each relationship, the source and target object indices are identified. If both objects meet distance constraints and the relationship probability exceeds a defined threshold, an edge is created between the corresponding nodes.
        
        \item \textbf{Inference improvement through Particle Filter (PF)}: 
        As the model's output is not accurate regarding mask inference, this leads to errors in calculating the object's centroid for obtaining its position relative to the map. A PF based on previous observations is applied to improve the accuracy of the result.
    \end{enumerate}
  \end{enumerate}
At the end of the process, the semantic map is updated with the new information, and the semantic scene is generated and provided to the planner module.
\\
The PF is used to track the object masks in real-time, provided as output by the PSGTR model, and to improve the estimation of their position in space.
During the update process, the filter uses information from frames acquired to refine the position estimate of the objects. The last object masks identified by the PSGTR model are compared with previous ones using the Intersection over Union (IoU) metrics and by applying the motion model, which can be defined as a transformation of the camera position relative to the map between two time instances. Denote the transformation matrices  describing the camera position at time \( t-1 \) and at subsequent time  \( t \),  \( \mathbf{T}_{t-1} \)  and \( \mathbf{T}_{t} \), respectively; then, the change in position and orientation can be expressed by the transformation matrix \( \Delta \mathbf{T} = \mathbf{T}_{t} \mathbf{T}_{t-1}^{-1} \). 
To associate objects between successive frames, we use an IoU matrix computed over segmentation masks. For two masks \( A \) and \( B \), IoU is defined as \( \text{IoU}(A, B) = \frac{|A \cap B|}{|A \cup B|} \), where \( |A \cap B| \) represents the area of intersection between masks \( A \) and \( B \), and \( |A \cup B| \) represents the area of their union.
To compare segmentation masks between two successive frames, we denote the segmentation mask at time \( t-1 \) as \( M_{t-1} \) and at time \( t \) as \( M_{t} \). The transformation matrix \( \Delta \mathbf{T} \) is applied to the previous mask to obtain a transformed mask \( M_{t-1}' \) such that \( M_{t-1}' = \Delta \mathbf{T} \cdot M_{t-1} \). The Intersection over Union (IoU) is then computed between the transformed mask \( M_{t-1}' \) and the current mask \( M_{t} \) as follows: \( \text{IoU}(M_{t-1}', M_{t}) = \frac{|M_{t-1}' \cap M_{t}|}{|M_{t-1}' \cup M_{t}|} \). This allows us to identify the same object across successive frames based on their masks.

More formally,  each object is represented by a set of \( N \) particles, where each particle \( s_i^t \) at time \( t \) is a 3D vector representing a hypothesis about the object's position: \( s_i^t = [x_i, y_i, z_i]^T \), where \( i = 1, \ldots, N \).
The particles are initialized with a normal distribution around the initially observed position \( \mu_0 = [x_0, y_0, z_0]^T \): \( s_i^0 \sim \mathcal{N}(\mu_0, \Sigma_0) \), where \( \Sigma_0 = \text{diag}(\sigma_x^2, \sigma_y^2, \sigma_z^2) \) is the initial covariance matrix.
Initial weights are uniform: \( w_i^0 = \frac{1}{N} \), where \( i = 1, \ldots, N \).
Prediction takes into account the camera motion. If \( T_{t-1,t} \) is the transformation matrix from frame \( t-1 \) to frame \( t \), each particle is updated as \( s_i^t = T_{t-1,t} \cdot s_i^{t-1} + s_i^0 \), where \( s_i^0  \) represents the noise added to account for uncertainties in motion, maintaining the same distribution structure used for initial particle initialization. Given a new observation \( s_{new} \), the particle weights are updated based on the Euclidean distance between the predicted position and the observed one: \( d_i^t = \|s_i^t - s_{new}\|_2 \) and \( w_i^t = \frac{1}{1 + d_i^t} \).
Weights are then normalized: \( w_i^t = \frac{w_i^t}{\sum_{j=1}^N w_j^t} \).
The final position of the object \( \hat{s}_t \) is estimated as the weighted mean of all the particles: \( \hat{s}_t = \sum_{i=1}^N w_i^t s_i^t \).
\\
Table \ref{tab:position_comparison} shows the improvement obtained over 30 measurements using particle filter.
\\
The overall process for updating the semantic map using the particle filter can be summarized by the algorithm \ref{alg:particle_filter_algorithm}.
\begin{algorithm}
\footnotesize
\caption{Semantic Map update using Particle Filter}
\begin{algorithmic}[1]
    \For{each frame $t$}
        \For{each object $k$}
            \State Apply transformation: $M_{t-1}' = \Delta \mathbf{T} \cdot M_{t-1}$  \Comment{Transform previous masks}
        \EndFor
        
        \State Compute $\text{IoU}(M_{t-1}', M_t) = \frac{|M_{t-1}' \cap M_t|}{|M_{t-1}' \cup M_t|}$ \Comment{Compute IoU between nodes and inference results}
        
        \For{each object $k$}
            \If{$\text{IoU} > \lambda_{\text{IoU}}$}
                \State Update weights: $d_i^t = \|s_i^t - s_{\text{new}}\|_2$, $w_i^t = \frac{1}{1 + d_i^t}$
                \State Normalize: $w_i^t = \frac{w_i^t}{\sum_{j=1}^{N} w_j^t}$
                \State Estimate: $\hat{s}_t = \sum_{i=1}^{N} w_i^t s_i^t$
            \EndIf
        \EndFor
        
        \For{each unmatched observation}
            \State Init new object: $s_i^0 \sim \mathcal{N}(\mu_0, \Sigma_0)$
        \EndFor
        
        \State Update semantic map with $\hat{s}_t$
    \EndFor
\end{algorithmic}
\label{alg:particle_filter_algorithm}
\end{algorithm}

\begin{table*}
\footnotesize
\caption{Comparison of position data}
\label{tab:position_comparison}
  \begin{tabular}{ccl}
    \toprule
   \textbf{Property} & \textbf{No Particle} & \textbf{Particle} \\
    \midrule
     Real position [m] & (0.67, 0.10, 0.95) & (0.67, 0.10, 0.95) \\
     Mean position [m] & (0.74, -0.08, 0.93) & (0.65, 0.08, 0.94) \\
     Mean of absolute error [m] & (0.07, 0.18, 0.02) & (0.02, 0.02, 0.01) \\
     Error standard deviation [m] & (0.35, 0.24, 0.03) & (0.17, 0.12. 0.02) \\
  \bottomrule%
\end{tabular}%
\end{table*}%

\section{Conclusions}
Scene graphs provide a structured representation that captures geometric and semantic information about the environment. This comprehensive understanding enables improved task planning with large language models, allowing robots to execute commands.

In this article we have shown how to use real-time sensor data  to dynamically update semantic maps, thus enabling the robot to adapt to ongoing changes in their environment, particularly in collaborative settings influenced by human actions. Here, particle filtering is applied to improve geometric data precision and semantic map accuracy. This can be particularly important also for social interaction and intention prediction \cite{ognibene2013contextual,rossi2017user} other than physical interaction with the environment.

The issues addressed in this work are cogent. Indeed, the effectiveness of planners in translating complex instructions into actionable plans relies on a robust state representation. Without an accurate semantic map, planners risk generating plans that misalign with the actual environment, potentially leading to task failures. The integration of semantic and geometric insights permits robots to reason about their environment in a more informed and adaptive way, ensuring that they can operate effectively and responsively in dynamic environments.

The adoption of a semantic map containing rich spatial information combined with a flexible LLM based planner can easily allow to explore in the future  the introduction of new spatial relationships, e.g. wrapped, stuck under, surrounding, aligned, that could support specific novel robot skills \cite{marocco2010grounding}.

\acknowledgments
Special thanks to Oversonic Robotics for enabling the implementation of the system using their humanoid robot, RoBee.

\bibliography{bibliography}

\section{Online Resources}
More information about RoBee and Oversonic Robotics are available:
\begin{itemize}
\item \href{https://oversonicrobotics.com/robee-humanoid-robot/?lang=en}{RoBee},
\item \href{https://oversonicrobotics.com/?lang=en}{Oversonic Robotics}
\end{itemize}
\end{document}